\pdfoutput=1

\documentclass[11pt]{article}

\usepackage{acl}

\usepackage{times}
\usepackage{latexsym}

\usepackage[T1]{fontenc}

\usepackage[utf8]{inputenc}

\usepackage{microtype}

\usepackage{mathtools}
\usepackage{tikz} 
\usepackage{tikz-qtree} 
\usepackage{tipa}
\usepackage{amsmath}
\usepackage{graphics}
\usepackage{tabularx}
\usepackage{xcolor}
\usepackage{multirow}
\usepackage{makecell}
\usepackage{placeins}

\usepackage{comment}
\usepackage[textsize=scriptsize]{todonotes}

\title{Analogy in Contact: Modeling Maltese Plural Inflection}

\author{Sara Court \\
  The Ohio State University \\
  {\tt court.22@osu.edu} \\\And
  Andrea D. Sims \\
  The Ohio State University \\
  {\tt sims.120@osu.edu} \\ \And
  Micha Elsner \\
  The Ohio State University \\
  {\tt elsner.14@osu.edu}}

\begin{document}
\maketitle
\begin{abstract}

Maltese is often described as having a hybrid morphological system resulting from extensive contact between Semitic and Romance language varieties. Such a designation reflects an etymological divide as much as it does a larger tradition in the literature to consider concatenative and non-concatenative morphological patterns as distinct in the language architecture. Using a combination of computational modeling and information theoretic methods, we quantify the extent to which the phonology and etymology of a Maltese singular noun may predict the morphological process (affixal vs. templatic) as well as the specific plural allomorph (affix or template) relating a singular noun to its associated plural form(s) in the lexicon. The results indicate phonological pressures shape the organization of the Maltese lexicon with predictive power that extends beyond that of a word's etymology, in line with analogical theories of language change in contact.
\end{abstract}

\section{Introduction}
Maltese is a Semitic language that has been shaped by an extensive history of contact with non-Semitic languages. A large influx of Sicilian, Italian, and English words over the course of hundreds of years has influenced the Maltese lexicon and grammar, making it a prime case study for those interested in the effects of language contact on morphological systems. Semitic languages are notable for their use of root-and-pattern (a.k.a. templatic) morphology in which inflectional or derivational forms of a lexeme may be related via the non-concatenative interleaving of consonants and vowels. In Maltese, some lexemes of non-Semitic origin have integrated into the native morphology to take both concatenative as well as non-concatenative patterns of Semitic origin. Non-Semitic morphological markers have also entered the grammar and may be found on lexemes of both non-Semitic and Semitic origin. 

This study applies methods from computational modeling and information theory to investigate factors shaping the organization of the modern Maltese lexicon.
Contextualized within frameworks of analogical classification and usage-based accounts of contact-induced language change, we quantify the extent to which the phonology and etymology of Maltese lexemes are predictive of nominal plural inflection in the language. 
The results indicate that system-level phonology, hypothesized to capture analogical pressures, and etymology, hypothesized to capture conservative pressures that resist analogical change, are predictive of Maltese plural inflection in non-redundant ways, with phonology being more predictive than etymology overall. 

Because Maltese is a Semitic language, we are also interested in the extent to which these factors are predictive of the type of morphology (either concatenative or non-concatenative) relating singular-plural pairs in the language. Our results show that both phonology and etymology are twice as predictive of a lexeme's plural allomorph(s) as compared to its concatenative type. This suggests that the analogical processes hypothesized to inform speakers' morphological intuitions are most sensitive to phonological similarities across surface forms, regardless of typological differences distinguishing concatenative and non-concatenative relationships. This study provides quantitative evidence for the role of analogical classification based on phonological similarity at the word level as a structuring principle of Maltese nominal plural morphology.

\section{Morphology in Contact: Maltese as a ``Hybrid'' Language?}

Maltese is a descendant of the Siculo Arabic variety spoken by settlers of the Maltese islands beginning in the year 1048 \cite{FabriMaltese,brincat2011maltese}. While the language is Semitic with respect to its genetic classification, isolation and centuries of foreign colonization led to the development of Maltese as a distinct language shaped by Sicilian, Italian, and English influence. Written records from as early as 1240 acknowledge Maltese as its own language \cite{brincat2017maltese}, but it was not until 1934 that Maltese was declared an official language of Malta, along with English and Italian \citep{FabriMaltese}. Italian was revoked as an official language in 1936, but its influence on the Maltese lexicon and grammar remains. 

Much of the existing literature on Maltese describes the language as having a ``split lexicon'' or a ``hybrid morphology'' \citep[e.g.,][]{spagnol2011tale,borg2017morphological}.
These characterizations reflect an etymological divide in the lexical stock. Semitic nouns in the language mostly form the plural with Semitic affixes or root-and-pattern templates, while non-Semitic nouns show a less strong tendency to form the plural with non-Semitic affixes. At the same time, hundreds of non-Semitic nouns inflect using Semitic patterns and are found
in nearly all plural classes \citep{borg_1997}. Integration in the opposite direction is also found for a smaller number Semitic nouns which
inflect using non-Semitic affixes. Maltese thus represents a partial, but not total, example of what has variously been called a ``stratal effect'' \citep{gardani2021spreads} or ``code compartmentalization'' \citep{friedman2013} or ``compartmentalized morphology'' \citep{matras2015}, in which native and borrowed morphological exponents in a language are restricted to applying to lexemes of the same etymological origin.

It is common in contact linguistics to describe outcomes of language contact as compositions of distinct linguistic systems, even in cases of extensive borrowing or codeswitching \citep[e.g.,][]{myers1997duelling,gardani2020borrowing}. Such descriptions are sometimes intended as theoretical analyses. For example, \citet{gardani2021spreads} treats the stratal effect not simply as an empirically observable pattern, but as a synchronic constraint within the grammar that is psychologically real for speakers: ``... a restriction on the application domain of non-native morphological formatives in a recipient language...'' \citep[132]{gardani2021spreads} that enforces the boundaries of etymologically-defined morphological subsystems.

However, we find the a priori assumption that stratal effects reflect distinct and psychologically real morphological subsystems to be problematic inasmuch as it conflates the property to be explained -- that language contact can result (to greater or lesser degree) in compartmentalized morphology -- with the mechanisms that produce and reinforce that compartmentalization. Stated differently, reification of the stratal effect as a mechanism of the grammar obscures important questions: Given that speakers do not generally know the etymological origins of words, how do they classify words into morphological patterns? What is the relationship between the processes that they use to do this and the stratal effect (or lack thereof) as an empirically observable outcome of language contact?

In this study we examine the (partial) stratal effect found in Maltese noun morphology, examining its relationship to factors known to be important outside of contact situations to how speakers classify words into morphological patterns. In particular we analyze the relative strength of a word's phonology and etymology as predictors of its nominal plural morphology and look at the relevance of these factors for the organization of the Maltese lexicon. It is important to note that we are not interested in etymology directly and we do not assume that speakers have or use direct knowledge of the etymology of words. We instead use etymology as a way to estimate the influence of conservative forces on morphological classification. We assume that the predictive power of etymology applies to words which have retained their etymological plurals, in some cases resisting pressures to conform to other parts of the language system. The conservative forces which resist these pressures include token frequency \citep{krause2022conservation}.

Additionally, as a related question, we ask whether there is evidence in Maltese for distinct morphological subsystems (``hybrid morphology'') in theoretical terms. This question is interesting in part because characterizations of Maltese as having hybrid morphology have also suggested, sometimes explicitly, that the non-concatenative morphology native to Semitic languages should be analyzed as distinct from concatenative morphology, both Semitic and non-Semitic. Moreover, research on morphological integration in Semitic languages has tended to focus specifically on the extent to which foreign words make use of native root-and-template morphology, as compared to affixation \citep[e.g.,][]{bensoukas2018,ziani2020}. However, since the vast majority of suffixal allomorphs in Maltese are of Semitic origin, division of the lexicon along etymological lines does not correspond to a split according to concatenative vs. non-concatenative morphology, as is sometimes implied. We test whether morphological type is a distinct factor in the stratal effect. Specifically, we ask whether there is support for analyzing root-and-pattern (templatic) plural morphology and affixal plural morphology as distinct subsystems.

We compare the results of two models: the first uses a lexeme's phonology and etymology to predict its concatenative type, either affixal or templatic. The second uses the same information to predict its inflectional allomorph, i.e., the specific affix or template found on the lexeme's plural form. Comparisons across factors within each model provide insight into the extent to which phonology and etymology are informative about plural morphology, and thus are likely to have played a role in the development of the language over centuries of contact with speakers of non-Semitic languages. Comparisons across the two models offer insight into the extent to which templatic and affixal morphological patterns operate as distinct subsystems in Maltese.

\section{Analogy and Language Change}

We take an analogical approach, using the term \textit{analogy} to refer broadly to any similarity-based, paradigmatic influence of one word on the morphological behavior of another. The importance of analogy as a mechanism of language change is well established in the field of historical linguistics \cite{anttila1977analogy,Hock+1991,fertig2013analogy, joseph2013phonically}, but it is most often discussed with respect to its role in language-internal change, independent of the effects of language contact. 
In contact linguistics, the idea that (phonologically-based) analogy plays a role in whether and how borrowed words are morphologically integrated into a recipient language has a long history, going back to at least \citet{haugen} and \citet{weinreich1953}. However, most analyses of lexical and morphological borrowing focus on the potential and observed outcomes of contact \cite[see][for an overview]{matras2020borrowing}, often with little to no discussion of the exact ways in which analogy is hypothesized to play a role. 

To examine the role of analogy, we take a cue from \citet{matras2009language}, who proposes a usage-based model of language contact in which a multilingual individual draws on a unified repertoire of linguistic resources. 
In this section we elaborate on how such a perspective can help in understanding the role of analogy, specifically analogical classification, 
in contact-induced morphological change and the development of the Maltese lexicon.

\subsection{The Paradigm Cell Filling Problem}\label{pcfp}

Analysis of the analogical mechanism hypothesized to drive morphological integration in contact may be understood as an extension of the Paradigm Cell Filling Problem (PCFP), a line of research in theoretical morphology that seeks to identify the information available to speakers that allows them to infer and produce grammatically inflected surface forms \cite{ackerman2009parts}. Most quantitative analyses of the PCFP to date take an analogical approach: speakers are hypothesized to rely on emergent similarities and paradigmatic relations among previously-acquired words in the lexicon to inform their intuitions when inflecting or processing rare or novel word forms \citep[see, e.g.,][]{ackerman2009parts,sims2016inflection,guzman2020analogy,parkerEtal2022}. 

\citeauthor{matras2009language}'s (\citeyear{matras2009language}) usage-based model of language contact is directly compatible with analogical approaches to the PCFP.
Since multilingual speakers are assumed to have access to a unified linguistic repertoire corresponding to all of their languages, this full repertoire may be drawn upon to make morphological generalizations. Combinations of generalizations from different languages during speech production may result in linguistic innovations or morphologically adapted ``nonce borrowings'' \cite{poplacketal1988}. 
Over time, some of these may be conventionalized and perpetuated throughout the larger speech community, leading to contact-induced language change.

We may therefore specify the PCFP with respect to language contact as follows: what guides speakers' grammatical intuitions when adapting and integrating lexemes in multilingual contexts, and how may conventionalized integration of borrowed linguistic material affect the intuitions of a monolingual speaker when producing inflected word forms?


\subsection{Computational Modeling of the PCFP}

A number of recent studies in computational linguistics have applied machine learning methods to analyze the kinds and amounts of information that may be available to speakers when solving the PCFP (in monolingual contexts).
For example, \citet{guzman2020analogy} uses a Long Short-term Memory Network \cite[LSTM,][]{hochreiter1996lstm} to quantify the respective informativity of stem phonology, lexical semantics, and affixal exponents as predictors of nominal inflection class organization in Russian. His results indicate that while each factor contributes predictive information, more information about inflection class is contributed by stem phonology than by any individual affix. Furthermore, the contributions of the three predictors are additive, indicating a level of nonredundancy in their informativity.

\citet{williams2020predicting} also employ the representational power of an LSTM to quantify the extent to which phonology and lexical semantics are predictive of a noun's declension class in German and Czech. As opposed to model accuracy, they measure the amount of Mutual Information, in bits, shared by phonology, semantics, and declension class systems in each language. They find that, while phonology is more predictive than semantics overall in both languages, the relative informativity of phonology and semantics varies greatly across the two languages and across individual declension classes within each language.

\citet{dawdy2014learnability} take an analogical approach to modeling plural formation in Modern Standard Arabic. The authors use a Generalized Context Model \cite[GCM,][]{nosofsky1990relations} to quantify the extent to which phonological factors, specifically similarities in consonant-vowel (CV) template (a.k.a. ``broken plural'' allomorph), segmental properties (in terms of natural classes), and lexical gang size \cite{alegre1999rule}, predict the form of a plural noun in Arabic. Their results indicate that all three factors are predictive to varying degrees, suggesting phonological representations that are both fine-grained, i.e., at the segmental level, and coarse-grained, i.e., with respect to gang size and CV template, may serve as a basis for analogical processing and morphological organization in Arabic.

Finally, \citet{nieder2021modelling, nieder2021knowledge} use both computational and psycholinguistic methods to investigate the role of analogical classification in the nominal plural system of Maltese. The authors find that plural forms in Maltese may be predicted with a reasonable degree of accuracy based on their phonological similarity to attested plural forms, modulated by the frequency distribution of plural allomorphs in the language. However, the authors do not specifically measure etymology as a predictor, leaving open the question of how non-Semitic words were integrated into the morphological system. In other words, it is unclear from their results whether phonology is predictive independently of etymology, 
or only as an indicator of etymological origin.

\section{Methods}

The current study adapts the methods proposed by \citet{williams2020predicting} to quantify the relative contributions of phonology and etymology as predictors of inflectional organization in Maltese. We use a character-level LSTM classifier trained to make inferences about a word's plural class by abstracting over the phonology of each word form as a whole. We then quantify the influence of phonology on Maltese nominal plural inflection using Mutual Information, an information theoretic measure of interpredictability among two or more systems. We compare our results to the predictive strength of the word's etymological origin using the same measures, quantifying the balance of analogical and conservative factors hypothesized to shape the integration of foreign lexemes into the grammar. 

\subsection{Data}
\label{sec:data}


This study merges data from two collections compiled by \citet{nieder2021osfwithout,nieder2021osfdiscriminative} into a single dataset consisting of 3,174 singular-plural noun pairs. 
Each pair is tagged for etymological origin, either Semitic or non-Semitic. The original data was manually compiled from the MLRS Korpus Malti v. 2.0 and 3.0 \cite{gatt2013digital} and supplemented with \citeauthor{schembri2012broken}'s (\citeyear{schembri2012broken}) collection of Maltese CV templates. Etymological information was sourced from a digitized version of \citeauthor{aquilina06}'s (\citeyear{aquilina06}) Maltese-English dictionary.
Plural nouns in the data are classified as taking one of 12 different suffixes (``sound plurals'') or 11 different non-concatenative CV templates (``broken plurals''), forming a nominal plural inflection system composed of 23 different inflection classes \cite{nieder2021osfwithout}. 
Maltese is the only standardized Semitic language written in a Latin script, using an orthography that ``represents the phonology of the language admirably'' according to \citet [258]{hoberman2007maltese}. For this reason, we analyze nouns using their original orthography, as in  \citet{williams2020predicting}.

Over 135 nouns in the dataset take more than one plural form. Of these, 78 nouns may take both broken and sound plurals. 
In this study, we account for these nouns by representing each pair separately at the allomorph level, whereas in the binary prediction model of the lexeme's concatenative type (concatenative vs. non-concatenative) we include a noun only once per type. For example, the word \textsc{libsa} `dress' may take the sound plural \textit{libsiet} and the broken plurals \textit{lbies} and \textit{lbiesi}. The lexeme \textsc{libsa} is therefore included in the model three times in the allomorph prediction setting, but only twice in the type prediction setting. 

Following \citet{williams2020predicting}, we remove all classes with fewer than 20 lexemes, leaving a total of 13 plural allomorph classes in our model. 
Table \ref{table:allomdist} shows the full distribution of allomorphs according to etymology and concatenative type. Note that lexemes that take more than one allomorph are counted more than once.

\begin{table}
\centering
\begin{tabular}{ccc||c}
\hline
& \textbf{Non-Semitic} & \textbf{Semitic} & \textbf{Total} \\
& \textbf{Lexeme} & \textbf{Lexeme} & \textbf{(\%)}\\
\hline
\textbf{Non-Semitic} & \multirow{2}*{1,274} & \multirow{2}*{21} & \multirow{2}*{42\%} \\
\textbf{Affix} & & & \\
\hline
\textbf{Semitic} & \multirow{2}*{416} & \multirow{2}*{684} & \multirow{2}*{35\%} \\
\textbf{Affix} & & & \\
\hline
\textbf{Semitic} & \multirow{2}*{240} & \multirow{2}*{537} & \multirow{2}*{23\%} \\
\textbf{Template} & & & \\
\hline
\hline
\multirow{2}*{\textbf{Total (\%)}} &  \multirow{2}*{62\%} & \multirow{2}*{38\%} & \multirow{2}*{100\%}\\
 & & & \\
\hline
\end{tabular}
\caption{Distribution of Maltese nominal plural allomorphs by lexeme etymology and concatenative type}
\label{table:allomdist}
\end{table}

\subsection{Formal Notation}
Following \citet{williams2020predicting}, we can define a lexeme as a tuple ($w_i, e_i, c_i)$ where for the $i^{th}$ lexeme, $w_i$ = the lexeme’s phonological form, $e_i$ = the lexeme’s etymological origin, and $c_i$ = the lexeme’s inflection class. We assume the lexemes follow a probability distribution $p(w,e,c)$, approximated by the corpus. We can define the space of $K$ inflection classes as $\mathcal{C}=\{1, ... ,K\}$, so that $c_i \in \mathcal{C}$ and define $C$ as the random variable associated with $\mathcal{C}$. For a set of lexemes derived from $N$ etymological origins, we can define an etymological space as $\mathcal{E} = \{1, ... , N\}$ so that $e_i \in \mathcal{E}$ and define $E$ as the random variable associated with $\mathcal{E}$. Each noun may be associated with one of two genders $g_i$ from the space of genders $\mathcal{G}$ specific to Maltese. Finally, we define the space of word forms as the Kleene closure over a language’s alphabet $\Sigma$, so that $w_i \in \Sigma$*, with $W$ as the random variable associated with $\Sigma$*. 

\subsection{Mutual Information (MI)}
Mutual Information ($\operatorname{MI}$) is an information theoretic measure that quantifies the degree of interpredictability among two or more systems.
%
For example, the $\operatorname{MI}$ shared by the nominal plural inflection class system $C$ and phonological system $W$ in  Maltese may be calculated as follows:

%
\begin{equation}
  \operatorname{MI}(C;W) = \operatorname{H}(C) - \operatorname{H}(C|W)  
  \label{eq:MI}
\end{equation}

This may be generalized to consider the amount of redundant information shared by inflection class, phonology, and etymology $E$ as follows:
\begin{equation}
    \operatorname{MI}(C;E;W) = \operatorname{MI}(C;W) - \operatorname{MI}(C;W|E)
    \label{eq:MI3}
\end{equation}

Because a language's grammatical gender system is known to interact with its inflectional morphology in non-deterministic ways \cite{corbett2000gender}, we follow \citet{williams2020predicting} and condition all relevant measures on gender:
\begin{equation}
    \operatorname{MI}(C;W|G) = \operatorname{H}(C|G) - \operatorname{H}(C|W,G)
    \label{eq:gender}
\end{equation}

\begin{figure}
    \centering
    \includegraphics[width=3.0truein]{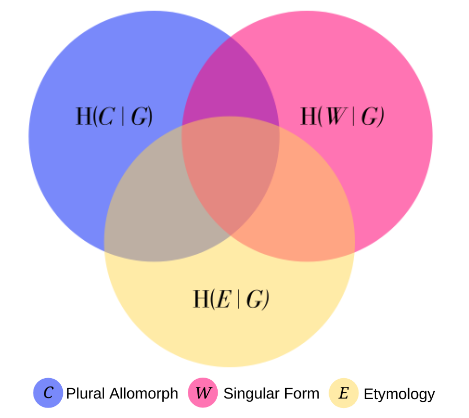}
    \caption{Tripartite Mutual Information}
    \label{fig:tripartiteMI}
\end{figure}

The intuitive reasoning behind Equations \ref{eq:MI} - \ref{eq:gender} may be seen in Figure \ref{fig:tripartiteMI}, in which each colored circle represents $\operatorname{H|G}$, the total entropy, conditioned on gender, of the three interacting systems under analysis.

Finally, since our corpus is only a sample of the language, we note that all calculations are estimates. However, while estimates over the finite inflection class and etymology systems can be empirically calculated using the corpus, the infinite number of possible word forms in the $\Sigma$* means calculations involving $W$ must be further approximated. Methods for estimating the entropy of both kinds of systems are described in detail in the following sections.

\subsection{Techniques for Estimating Entropy}
We use plug-in estimation to obtain entropy values for $C$ and $E$, calculating the distribution $p(c)$ for $c \in C$ (or alternatively, $p(e)$ for $e \in E)$ and using this to estimate $\operatorname{H}(C)$ in Equation \ref{eq:MI} above. 
\subsection{Approximating Conditional Entropy}
 $\operatorname{H}(C|E)$ may be similarly calculated using plug-in estimation. 
 However, given the infinite number of possible word forms in $\Sigma$*, an estimate for $\operatorname{H}(C|W)$ cannot be calculated directly from the corpus. We therefore approximate this value using cross-entropy, which has been mathematically proven to be an upper bound on conditional entropy \cite{brown1992estimate}.
%
%
%
We use the cross-entropy loss obtained from a computational model that has been trained to predict the plural class $c_i$ associated with a singular noun $w_i$ to approximate the cross-entropy of the system:

\begin{equation}\label{eq:crossentavg}
    \operatorname{H}(C|W) \le  -\frac{1}{M}\sum_{i=1}^{M} \log q(c_i|w_i)
\end{equation}

We note that as the amount of data in the corpus increases, i.e., as $M \to \infty$, the above value approaches the true cross-entropy value.

\subsection{Normalized Mutual Information (NMI)}
To compare results across models and across languages, we normalize MI values by dividing by the total entropy of the inflection class system. For example, the $\operatorname{NMI}$ shared by a Maltese noun's phonology and plural inflection may be calculated as:
\begin{equation}
    \operatorname{NMI}(C;W) = \frac{\operatorname{MI}(C;W)}{\operatorname{H}(C)}
\end{equation}

\subsection{Model Details}
We adapt the LSTM classifier implemented in \citet{williams2020predicting} to estimate the probability that a plural class $c$ is associated with a given input noun $w$ of gender $g$, i.e., $q(c|w,g)$ in Equation \ref{eq:crossentavg}. The model learns a set of character embeddings to represent the phonological forms of singular nouns as part of the training process. Gender is separately embedded and input into the model's initial hidden state. The model is trained using Adam \cite{kingma2014adam} with model hyperparameters, including the number of training epochs and the number and sizes of hidden layers, optimized using the Bayesian optimization technique implemented in \citet{williams2020predicting}. The model then learns a probability distribution that serves to approximate $q(c|w,g)$. 

Following training, we test the model on a held-out dataset and use the model's cross-entropy loss to serve as an approximate upper bound on the conditional entropy $\operatorname{H}(C|W,G)$. We use 10-fold cross validation to make full use of the dataset for our approximations. To estimate $q(c|w,e,g)$, we concatenate a binary character representing the word's etymology onto the end of the noun to serve as model input and follow the same procedure.

\section{Results}

$\operatorname{NMI}$ and $\operatorname{H}(C|G)$ values for $C$ defined as concatenative type and plural allomorph, respectively, are presented in Table \ref{table:NMI}. 
%
The largest $\operatorname{NMI}$ value we obtain, $\operatorname{NMI}(E;W|G)$, indicates that more than half of the information needed to predict a word's etymology is shared with its phonology. In other words, it is often not difficult to guess the origin of a Maltese word based on how it sounds. 
Note that this value is consistent across models, as it does not depend on $C$. 

\begin{table}
\begin{center}
\begin{tabular}{lccc}
\hline
&& \textsc{\textbf{TYPE}} & \textsc{\textbf{ALLO.}} \\
\hline
$\operatorname{H}(C|G)$& 
    \begin{minipage}{.4cm}
      \includegraphics[width=\linewidth]{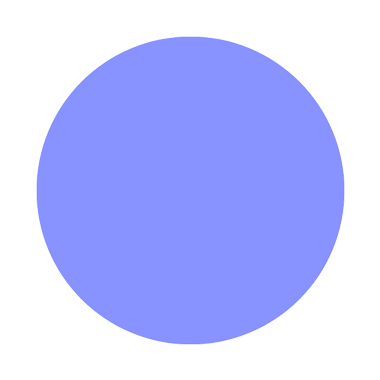}
    \end{minipage}
&0.81 & 2.65\\
$\operatorname{NMI}(C;W|G)$ &
    \begin{minipage}{.4cm}
      \includegraphics[width=\linewidth]{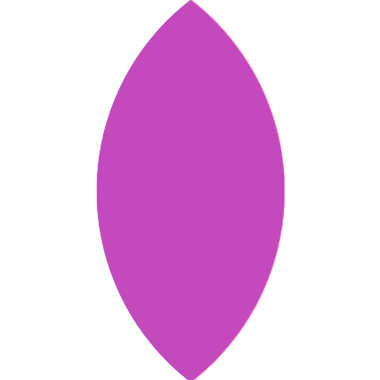}
    \end{minipage}
    & 0.21 & 0.42 \\
$\operatorname{NMI}(C;E|G)$ &
    \begin{minipage}{.3cm}
      \includegraphics[width=\linewidth]{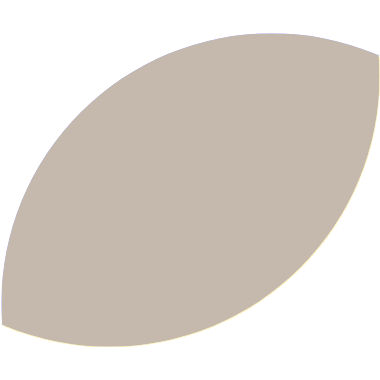}
    \end{minipage}
    & 0.13 & 0.22 \\
$\operatorname{NMI}(C;E;W|G)$ &
    \begin{minipage}{.4cm}
      \includegraphics[width=\linewidth]{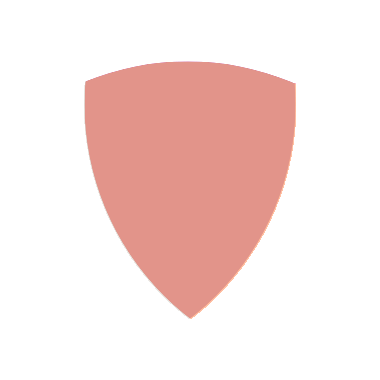}
    \end{minipage}
    & 0.06 & 0.15\\
$\operatorname{NMI}(E;W|G)$ &
    \begin{minipage}{.3cm}
      \includegraphics[width=\linewidth]{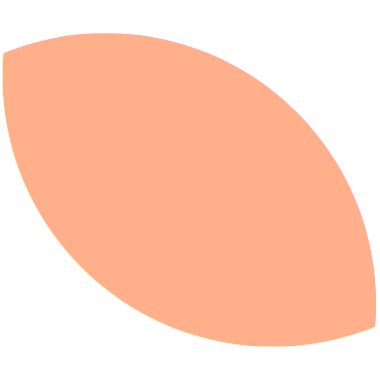}
    \end{minipage}
    &0.61 & 0.61 \\
\hline
\end{tabular}

\caption{\label{table:NMI}Normalized Mutual Information measures for plural class $C$ defined with respect to \textsc{type} vs. \textsc{allomorph}. NMI values involving $C$ are normalized with respect to $\operatorname{H}(C|G)$, while $\operatorname{NMI}(E;W|G)$ is normalized with respect to $\operatorname{H}(E|G)$.}
\end{center}
\end{table}

\subsection{Concatenative Type}
Results for the model predicting a noun's concatenative type
are in Table \ref{table:NMI}. Note first that the entropy $\operatorname{H}(C|G)$ of the plural inflection class system defined at the level of concatenative type is calculated to be 0.81, indicating that, given its gender, predicting whether a random Maltese noun takes concatenative or non-concatenative morphology
is more predictable than chance, although not by much. We find phonology, indicated by $\operatorname{NMI}(C;W|G)$, to be more predictive than etymology, indicated by $\operatorname{NMI}(C;E|G)$.
Crucially, each of these bipartite NMI values exceeds the tripartite mutual information $\operatorname{NMI}(C;E;W|G)$ shared across all three systems. This indicates that while a non-trivial amount of predictive information is shared across all three systems, phonology and etymology are each predictive of concatenative type in partially non-redundant ways. 
This suggests that both analogical and conservative forces are likely to have played a role in the development of the Maltese nominal plural system. 

\subsection{Plural Allomorph}
In an analogical model of inflection in which singular inflected forms and their plural counterparts share a direct relationship in the lexicon, the predictive principles structuring the morphological system are expected to be most evident when defining an inflection class system at the level of the allomorph.

We first note that the entropy $\operatorname{H}(C|G)$ calculated over the plural class distribution defined at the allomorph level is nearly three times as high as the entropy of $C$ when defined as a noun's concatenative type. This is reflective of the higher degree of unpredictability associated with a non-uniform distribution of nouns over a greater number of inflection classes. When comparing across the allomorph and concatenative type models it is thus important to normalize for the fact that predicting allomorphs is more difficult than predicting concatenative type. However, even calculations normalized in this way show that the interpredictability among phonology, etymology, and plural inflection, indicated by the NMI values in Table \ref{table:NMI}, are all twice as high at the allomorph level as they are for concatenative type. In other words, a noun's singular form reduces the relative uncertainty about its plural allomorph twice as much as it reduces the uncertainty about whether that allomorph is concatenative.
This suggests the analogical and conservative pressures hypothesized to shape morphological organization are more sensitive to correspondences at the word level than to typological similarities with respect to concatenativity. 

Additionally, the general tendency found at the level of concatenative type still follows when classes are defined at the level of individual allomorphs: phonology shares more information with inflection class than does etymology, with each factor contributing some amount of non-redundant information. This illustrates one key advantage of the methods employed in this study, namely the ability to disentangle the independent contributions of either predictor from the degree to which both exert redundant organizational pressure towards the same end. 

For example, given the fact that phonology and etymology are themselves mutually informative, we cannot uniquely interpret either bipartite measure of $\operatorname{MI}$, that is, $\operatorname{NMI}(C;W|G)$ or $\operatorname{NMI}(C;E|G)$, as indicative of the forces hypothesized to shape the integration of linguistic material in contact. Rather, evidence for analogical structuring of the Maltese plural system at the allomorph level is specifically indicated by the positive difference between $\operatorname{NMI}(C;W|G)$ and $\operatorname{NMI}(C;E;W|G)$. Conservative pressures, such as those associated with high token-frequency items \cite{krause2022conservation}, are similarly indicated by the extent to which $\operatorname{NMI}(C;E|G)$ exceeds $\operatorname{NMI}(C;E;W|G)$.

\subsection{Variation Across Allomorph Classes}

\begin{figure}[!htbp]
    \centering
    \includegraphics[width=3.0truein]{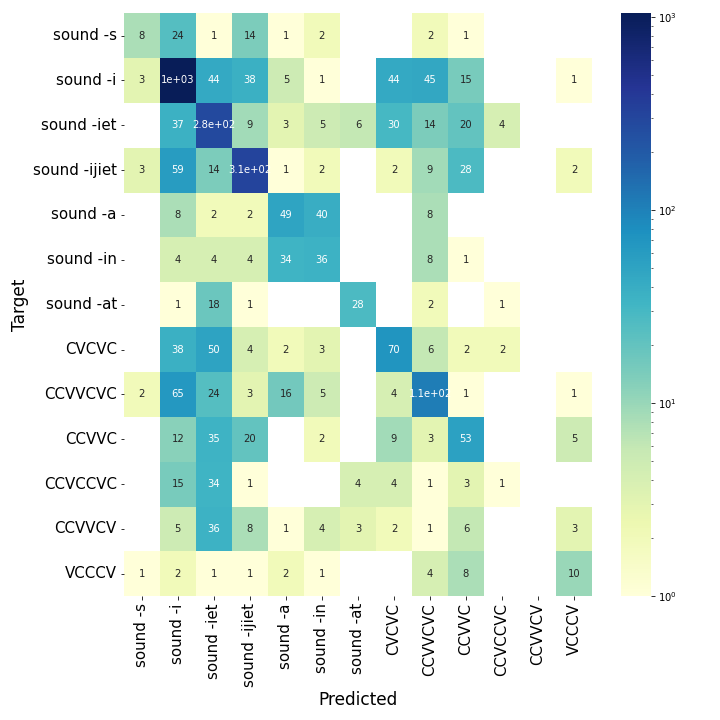}
    \caption{Confusion matrix: predicting plural allomorph from singular phonology and gender}
    \label{fig:confmatrix}
\end{figure}

Closer examination of the model's predictions reveals an effect of type frequency, with larger inflection classes 
predicted more often than smaller classes. Table \ref{table:modelacc} reports the accuracy of all models in which singular noun phonology $W$ is a predictor. Since all models achieve an overall accuracy above a majority baseline, the $\operatorname{NMI}$ values we obtain may be reliably interpreted as empirical minimums. However, as can be seen in Figure \ref{fig:confmatrix}, the model's incorrect predictions do not clearly distinguish between sound and broken classes; nouns with a sound plural allomorph may be misclassified as taking a broken plural template, and nouns taking a broken plural may be incorrectly predicted to take a sound plural.

\begin{table}[!htbp]
    \centering
    \begin{tabular}{W{c}{2.5cm}W{c}{2.7cm}W{c}{1.2cm}}
         \textbf{Target} & \textbf{Model} & \textbf{Accuracy} \\
         \hline
         \multirow{2}*{\textsc{\textbf{Etym. ($E$)}}} &$\operatorname{MI}(E;W|G)$ & 0.90 \\
         & Baseline & 0.62 \\
         \hline
         \multirow{3}*{\textsc{\textbf{Type ($C$)}}} & $\operatorname{MI}(C;W|G)$ & 0.80\\
         &$\operatorname{MI}(C;E;W|G)$ & 0.81 \\
         & Baseline & 0.77 \\
         \hline
         & $\operatorname{MI}(C;W|G)$ & 0.65\\
         \textsc{\textbf{Allomorph}}& $\operatorname{MI}(C;E;W|G)$ & 0.68 \\
         ($C$)& Baseline & 0.40 \\
    \end{tabular}
    \caption{Model accuracy for all models predicting Etymology $E$ or Plural Class $C$ (Type vs. Allomorph) using the Phonology $W$ of singular nouns in Maltese}
    \label{table:modelacc}
\end{table}

%


If speakers are sensitive to differences between concatenative and non-concatenative allomorphs grouped into high-level macro classes (morphological subsystems), we might expect some degree of observable within-class coherence with respect to either or both of the phonology and etymology of words exhibiting a particular morphological behavior. Specifically, we would expect a pattern of predictions in which the LSTM is able to first identify a lexeme's concatenative type before predicting, possibly incorrectly, an allomorph of that specific type. 

Instead, as seen in Figure \ref{fig:confmatrix}, we do not find such evidence.
Rather, we find evidence for coherence at the allomorph level, specifically, for phonological patterns as a predictor of inflectional organization and driver of inflectional behavior at the allomorph level. 

 

\begin{figure}[!htbp]
    \centering
    \includegraphics[width=3.0truein]{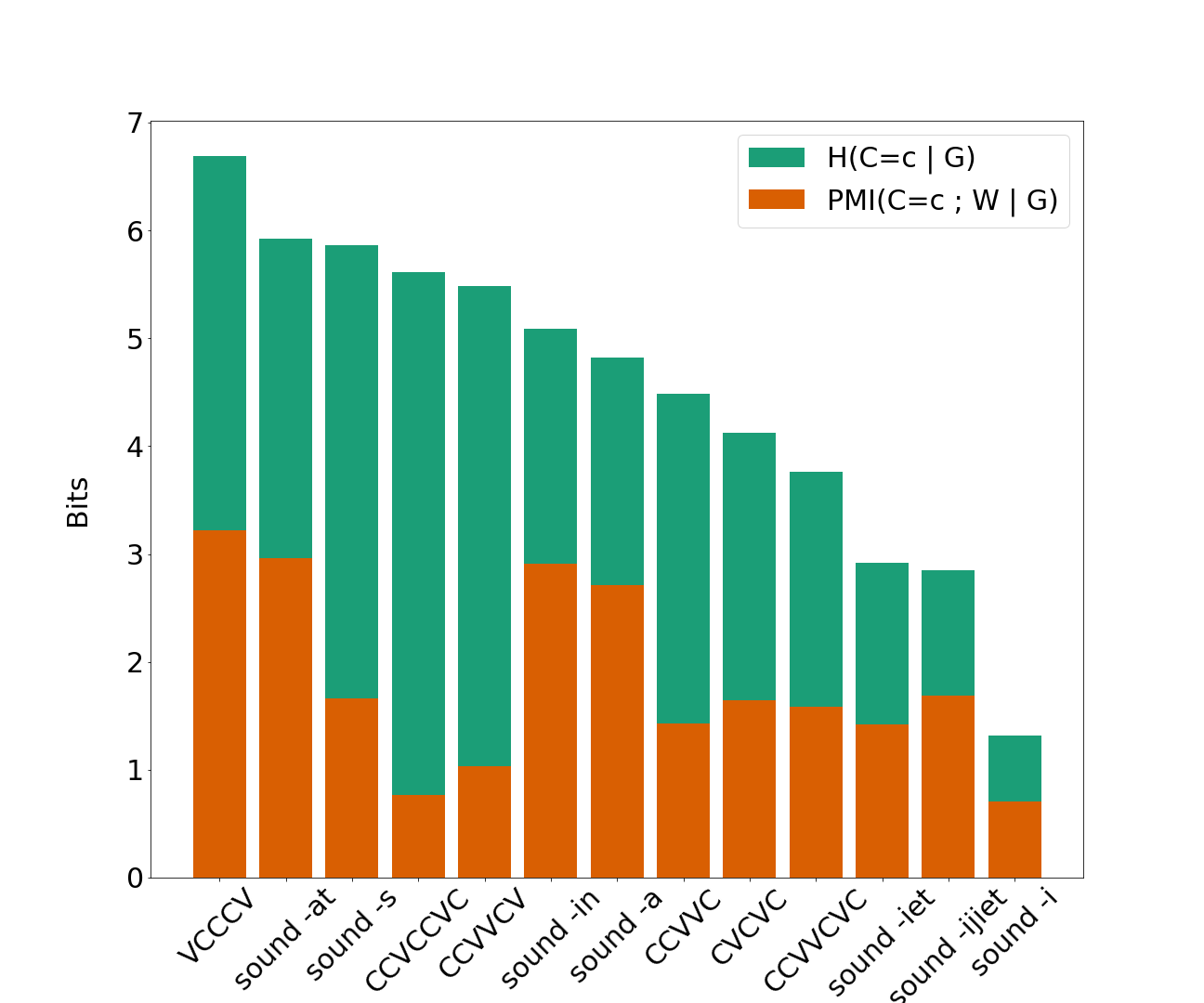}
    \caption{Partial Pointwise Mutual Information $(\operatorname{PMI})$ shared by word form and class for each allomorph class}
    \label{fig:pmi}
\end{figure}

Finally, as in \citet{williams2020predicting}, we also conduct an analysis of the partial Pointwise Mutual Information $(\operatorname{PMI}$) shared between phonology $W$ and class $C$ with respect to the surprisal $\operatorname{H}(C=c|G)$ for each class, defined at the allomorph level. Figure \ref{fig:pmi} shows this distribution, with allomorph classes presented in order of increasing type frequency (and thus decreasing surprisal). We note that Maltese noun classes are each only partially predictable given the phonology of words belonging to them, regardless of class size or etymological origin. 

\section{Discussion}

In this paper we used an LSTM to help estimate the kinds and amounts of information that may be available to speakers when ``solving'' the PCFP. Overall, our results provide quantitative evidence for the role of both word phonology and etymology (as a stand-in for conservative factors) in shaping the Maltese lexicon.

Specifically, we found that the extent to which a Maltese singular noun's phonology predicts its plural morphology exceeds that of etymology in non-redundant ways. This suggests that analogical pressures from phonological correspondences across the lexicon shape nominal plural inflection in Maltese, independently of the etymological source language for some word or morphological pattern. 

Our results also show an independent contribution of etymology as a predictor.
We hypothesize that this captures conservative pressures theorized to resist analogical change, including token frequency \cite{krause2022conservation}. It may also reflect associative correlations from the use of lexemes of a common etymology in similar contexts, strengthening their coherence as a subsystem in the multilingual repertoire and encouraging the maintenance of a noun's original morphology. Further work is needed to investigate these possibilities.


In language contact situations such as that of Maltese, it is likely that an influx of foreign lexemes and increased productivity of foreign affixes affect both the size and character (e.g., phonology) of nominal plural classes relative to each other over time. This in turn is likely to affect subsequent classification and integration of words into the inflectional morphology of the language. 

In general, our results do not support characterizations of Maltese in which concatenative and non-concatenative morphologies co-exist as discrete systems within the lexicon. While a singular noun's phonology and etymology are each somewhat predictive of its concatenative type, they are twice as predictive of the actual plural allomorph(s) with which the lexeme is associated. This suggests that systematic relationships at the word level organize the morphology of Maltese, in turn shaping the language as new words are integrated and inflected.

\section{Conclusion}
This study extends previous work in information theory, computational modeling, and theoretical morphology to provide quantitative evidence for the role of phonology as an analogical force in the morphological organization of Maltese.
We ground this in a usage-based account of multilingualism and contact-induced change in which speakers are hypothesized to make use of analogical reasoning, among other language-general cognitive functions, when integrating novel words and patterns within a unified linguistic repertoire. The same processes that guide synchronic language use are proposed to be responsible for the diachronic effects of contact-induced language change. Specifically, it is hypothesized that speakers draw on similarities across multiple dimensions -- including but not limited to phonological patterns, semantic and indexical meaning, pragmatic function, and contexts of use -- to collaboratively construct and adapt grammatical systems of linguistic communication over time.

In the case of Maltese, our findings indicate that while a lexeme's phonology and etymology are themselves highly interpredictable, each contributes non-redundant information to reduce uncertainty when predicting the lexeme's plural inflection. While the etymology of a noun is somewhat predictive of its plural inflection, the word's phonology plays a much greater role. This synchronic analysis has diachronic implications. Our results suggest that analogical pressures from phonological similarities across the lexicon may have guided speakers' inflectional behavior when code mixing over the course of the development of the language to result in the conventionalized forms observed in modern Maltese. However, further diachronic study is needed to confirm this interpretation.

Contrary to a hypothesis in which concatenative and non-concatenative systems operate as separate subsystems within a ``split'' or ``hybrid'' morphology, our results indicate correspondences at the level of individual wordforms and affixes are driving speakers' morphological behavior. Specifically, the phonology and etymology of a lexeme are twice as predictive of its plural allomorph than its concatenative type. Further investigation into Maltese nouns attested to take plural forms of both concatenative types may provide additional insight into the ways in which concatenative type affects speakers' behavior, if at all. Future work should also consider additional factors known to shape inflection class systems, for example by integrating semantic word vectors into the model. Finally, additional comparisons implementing these methods across corpora in a variety of languages will continue to shed light on the factors shaping morphological systems cross-linguistically.

\section*{Acknowledgments}
We thank Jessica Nieder and Adam Ussishkin for generously sharing an abundance of digital resources and helpful feedback, Sarah Caruana for her insight into the Maltese language, and Christian Clark and Andrew Duffy for their contributions to an initial version of this project.

This material is based on work supported by the National Science Foundation under grant BCS-2217554 (\textit{Neural discovery of abstract inflectional structure}, PI Micha Elsner, Co-PI Andrea Sims).

\bibliography{anthology,custom}
\bibliographystyle{acl_natbib}

\newpage
\onecolumn
\appendix

\section{Nominal Plural Allomorphs in Maltese}
\FloatBarrier
\label{sec:appendix}
\begin{center}
  

\bigskip

\begin{tabular}{W{l}{1.3cm}lll}
\textbf{Sound Plural}\\
\hline
    Singular & Plural & Gloss & Allomorph\\
\hline
    \textit{karta} & \textit{karti} & `paper' & -\textit{i}\\
    \textit{omm} & \textit{ommijiet} & `mother' & -\textit{ijiet}\\
    \textit{rixa} & \textit{rixiet} & `feather' & -\textit{iet}\\
    \textit{giddieb} & \textit{giddieba} & `liar' & -\textit{a}\\
    \textit{me\textcrh lus} & \textit{me\textcrh lusin} & `freed' & -\textit{in}\\
    \textit{kuxin} & \textit{kuxins} & `cushions' & -\textit{s}\\
    \textit{triq} & \textit{triqat} & `street' & -\textit{at}\\
    \textit{sid} & \textit{sidien} & `owner' & -\textit{ien}\\
    \textit{ba\textcrh ri} & \textit{ba\textcrh rin} & `sailor' & -\textit{n}\\
    \textit{\textcrh ati} & \textit{\textcrh atjin} & `guilty' & -\textit{jin}\\
    \textit{spalla} & \textit{spallejn} & `shoulder' & -\textit{ejn}\\
    \textit{sieq} & \textit{saqajn} & `foot' & -\textit{ajn}\\
    \textit{qieg\textcrh } & \textit{qieg\textcrh an} & `bottom' & -\textit{an}\\
\end{tabular}
\captionof{table}{Sound plural allomorphs in Maltese, from \citet{nieder2021osfwithout}}
\label{tab:soundplurals}

\end{center}

\bigskip


\begin{center}

\begin{tabular}{W{l}{1.3cm}lll}
\textbf{Broken Plural}\\
\hline
    Singular & Plural & Gloss & Allomorph\\
\hline
    \textit{fardal} & \textit{fradal} & `apron' & CCVVCVC\\
    \textit{birra} & \textit{birer} & `beer' & (C)CVCVC\\
    \textit{kbir} & \textit{kbar} & `big' & CCVVC\\
    \textit{ftira} & \textit{ftajjar} & `type of bread' & CCVjjVC\\
    \textit{bit\textcrh a} & \textit{btie\textcrh i} & `yard' & CCVVCV\\
    \textit{sider} & \textit{isdra} & `chest' & VCCCV\\
    \textit{marid} & \textit{morda} & `sick person' & CVCCV\\
    \textit{g\textcrh odda} & \textit{g\textcrh odod} & `tool' & (g\textcrh )VCVC\\
    \textit{elf} & \textit{eluf} & `thousand' & VCVC\\
    \textit{g\textcrh aref} & \textit{g\textcrh orrief} & `wise man' & CVCCVVC(V)\\
    \textit{g\textcrh ama} & \textit{g\textcrh omja} & `blind person' & (g\textcrh )VCCV\\
\end{tabular}
\captionof{table}{Broken plural allomorphs in Maltese, from \citet{nieder2021osfwithout}}
\end{center}

\end{document}